\documentclass[conference]{IEEEtran}
\IEEEoverridecommandlockouts

\usepackage{cite}
\usepackage{amsmath,amssymb,amsfonts}
\usepackage{graphicx}
\usepackage{textcomp}
\usepackage{xcolor}
\usepackage{booktabs}
\usepackage{url}
\usepackage{algorithm}
\usepackage{algpseudocode}
\algrenewcommand\algorithmicrequire{\textbf{Input:}}
\algrenewcommand\algorithmicensure{\textbf{Output:}}

\def\BibTeX{{\rm B\kern-.05em{\sc i\kern-.025em b}\kern-.08em
    T\kern-.1667em\lower.7ex\hbox{E}\kern-.125emX}}

\begin{document}

\title{FETALFusion: Resolution-Aware State-Space Modeling for
Multi-Domain Fetal Head Segmentation}

\author{
\IEEEauthorblockN{Ammar Bhilwarawala}
\IEEEauthorblockA{
  \textit{School of Computer Engineering,}\\
  \textit{KIIT Deemed to be University}\\
  Bhubaneswar, India\\
  ammarbhilwarawala@gmail.com}
\and
\IEEEauthorblockN{Mainak Bandyopadhyay}
\IEEEauthorblockA{
  \textit{School of Computer Engineering,}\\
  \textit{KIIT Deemed to be University}\\
  Bhubaneswar, India\\
  mainak.bandyopadhyayfcs@kiit.ac.in}
}

\maketitle

\begin{abstract}
Head circumference (HC) measured from 2D ultrasound is a routine marker of
fetal growth, and automating it needs segmentations whose boundaries hold up
on individual scans, not only on average. We present FETALFusion, a dual-path
CNN-Mamba encoder-decoder for fetal head segmentation across acquisition
types. It applies four-directional state-space scans only at the two
highest-resolution stages, routes bottleneck features through domain-specific
channel gates that are averaged at test time, and adds a weak landmark shape
prior. We train jointly on transabdominal (HC-18) and transperineal (PSFHS)
images and compare ten methods on a held-out test set of 471 images, each
trained under one schedule and scored by surface distances at native
resolution. FETALFusion reaches a Dice score of 0.963 and an HD95 of
2.10\,mm without test-time augmentation or a domain label. It improves on
U-Mamba on all five reported metrics (HD95 reduced by 0.31\,mm;
Holm-corrected $p<10^{-4}$), a lead that holds under test-time augmentation
and across two independent training runs, and in the mean on U-Net,
Attention U-Net and TransUNet. Residual encoders close the gap: a ResNet-34
UNet++ exceeds FETALFusion, and an ImageNet-initialised ResNet-34 U-Net is
best overall (HD95 1.75\,mm). The ablation agrees. Calibrated against
run-to-run variation, it attributes FETALFusion's measurable gains to the
landmark shape prior and, on transperineal images, to the domain gates, while
removing the state-space branch does not measurably change accuracy. HC from
the predicted masks agrees with the clinical reference at an intraclass
correlation of 0.9986.
\end{abstract}

\begin{IEEEkeywords}
Fetal head segmentation, state-space models, multi-domain generalization,
ultrasound image analysis, boundary-aware segmentation.
\end{IEEEkeywords}

\begin{figure*}[t]
  \centering
  \includegraphics[width=\textwidth]{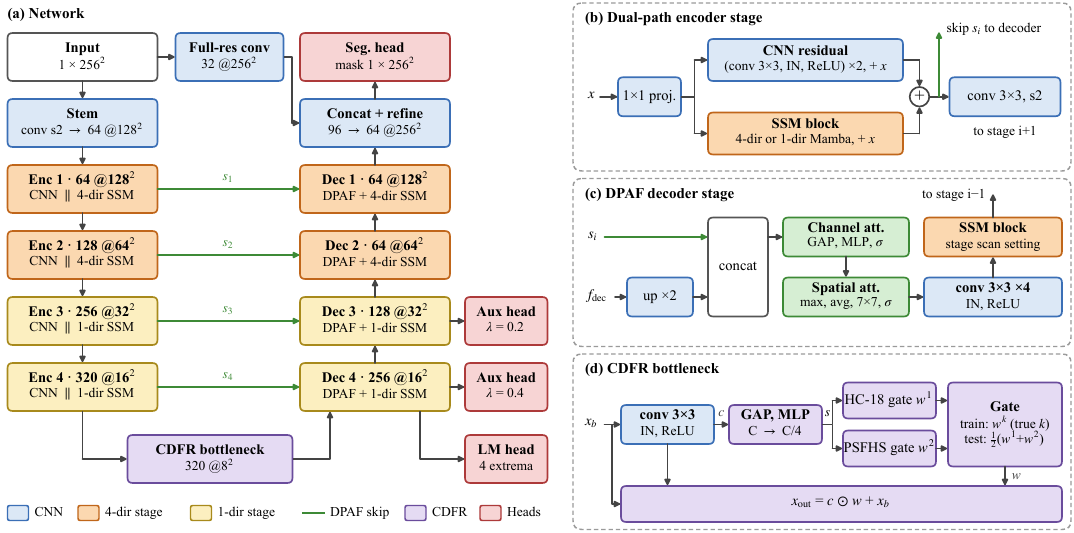}
  \caption{FETALFusion. (a) U-shaped network with four-directional state-space
    scanning at $128^2$ and $64^2$ and single-directional scanning at $32^2$ and
    $16^2$, in both encoder and decoder. Skips $s_i$ enter the decoder through
    DPAF. The auxiliary heads on decoder stages 4 and 3 and the landmark head on
    stage 4 are used only during training. (b) Encoder stage: a residual CNN
    branch and a state-space branch process the same input and are summed
    (Eq.~\ref{eq:fused}); the $1{\times}1$ projection is the identity when
    channel counts match. (c) Decoder stage: DPAF fuses the upsampled decoder
    features with the skip, followed by convolutions and a state-space block.
    (d) CDFR: the gate is selected by the true acquisition type during training
    and averaged at inference, so no domain label is needed.}
  \label{fig:arch}
\end{figure*}

\section{Introduction}

Fetal head circumference (HC) measured from 2D ultrasound is used to track
growth through pregnancy and to flag microcephaly and
macrocephaly~\cite{hc18}. Manual measurement varies between operators, and
automated segmentation is the obvious route to standardised biometry. The
harder part is making it dependable on the individual scan: a method that is
excellent on average can still return a badly wrong contour for a particular
patient.

The difficulty is geometric. The fetal skull appears as a thin elliptical rim
whose opposite sides can sit a large fraction of the image apart. A
convolutional encoder relates the two sides only after heavy downsampling, by
which point the rim has blurred. Our earlier Attention-ResUNet~\cite{bhilwarawala2025attres}
strengthened a residual U-Net for this task with multi-scale attention gates,
but it inherits the same limit. Self-attention covers that range directly,
as in TransUNet~\cite{chen2021transunet}, but its cost grows quadratically
with the number of tokens, which is expensive at the resolutions where the
boundary is still sharp.

State-space models offer linear-time sequence modelling with input-dependent
gating. Mamba~\cite{gu2024mamba} has been adapted to medical segmentation by
flattening 2D feature maps into 1D sequences~\cite{ma2024umamba,ruan2024vmunet},
and later work refines the serialisation itself, with high-order interactions
in the 2D selective scan~\cite{wu2025hvmunet} or token clustering for local
context~\cite{zhu2025mcc}. Row-major flattening is weak at high resolution:
two boundary points at the top and bottom of the skull, a few dozen pixels
apart in the image, sit several thousand positions apart in the sequence at
$128^2$, and a causal scan carries information one way only. Column-wise and
reversed scans address both. At $32^2$ the whole map is 1,024 tokens, and the
extra directions have far less to add.

That asymmetry suggests assigning scan directions by resolution.
FETALFusion (Fig.~\ref{fig:arch}) is built around the idea, with a mechanism
for the domain shift between transabdominal and transperineal acquisitions
that needs no scanner label at test time. We evaluate it under a strict
protocol: a held-out test split, one training schedule for every model,
surface distances at native resolution, and an ablation read against
run-to-run variation.

\noindent\textbf{Contributions.}
\begin{enumerate}
\item \textbf{FETALFusion}, a dual-path CNN-Mamba segmenter with
  resolution-aware scanning, Conditional Domain Feature Routing (CDFR) and a
  landmark shape prior. Under matched training it improves on U-Mamba and VM-UNet on every reported
  metric and, in the mean, on U-Net, Attention U-Net, an nnU-Net-style
  ResU-Net and TransUNet, with and without test-time augmentation; its HC
  estimates agree with the clinical reference at ICC 0.9986.
\item \textbf{Label-free multi-domain inference.} Averaging the domain gates
  at test time loses nothing against routing with the true domain label, and
  on transperineal images the gates improve boundary accuracy over a single
  shared gate.
\item \textbf{A calibrated study.} Every ablation effect must exceed
  run-to-run variation measured over two training runs, and ten baselines
  separate the encoder and its initialisation from the state-space design.
  The landmark prior, domain gating on the harder domain and the placement of
  four-directional scans matter; at this data scale the state-space branch as
  a whole and the attention fusion do not.
\end{enumerate}

\section{Method}

\subsection{Resolution-Aware Dual-Path Encoder}

FETALFusion follows the U-shaped layout of U-Net~\cite{ronneberger2015unet}.
A stride-2 convolutional stem maps the $256^2$ input to $128^2$ at 64
channels. Instance Normalization~\cite{ulyanov2016instance} is used
throughout, as it is less sensitive to intensity differences between probes
and vendors~\cite{isensee2021nnu}, and a two-layer projection keeps a
32-channel full-resolution map, concatenated with the last decoder output so
the thin rim survives downsampling.

Four encoder stages with channels $[64,128,256,320]$ emit skip connections at
$128^2$, $64^2$, $32^2$ and $16^2$, with a bottleneck at $8{\times}8$. Each
stage runs two branches on the same input and sums them:
\begin{equation}
  \mathbf{f} = \text{CNN}_\text{res}(\mathbf{x}) + \text{Mamba}_\text{RA}(\mathbf{x}),
  \label{eq:fused}
\end{equation}
where $\text{CNN}_\text{res}$ is a two-layer residual convolutional block and
$\text{Mamba}_\text{RA}$ is the resolution-aware state-space block.

\noindent\textbf{Resolution-aware scanning.}
At $128^2$ and $64^2$, four Mamba instances (state dimension 16, convolution
width 4, expansion factor 2) scan the flattened map left-to-right,
right-to-left, top-to-bottom and bottom-to-top, and a learned projection
merges them:
\begin{equation}
  \mathbf{y}
    = W_\text{m}\!\bigl[
        \text{S}_\text{lr}(\mathbf{z});\,
        \text{S}_\text{rl}(\overleftarrow{\mathbf{z}});\,
        \text{S}_\text{tb}(\mathbf{z}_\text{c});\,
        \text{S}_\text{bt}(\overleftarrow{\mathbf{z}_\text{c}})
      \bigr],
  \label{eq:4dir}
\end{equation}
with $\mathbf{z}\!\in\!\mathbb{R}^{HW\times d}$ the row-major sequence,
$\mathbf{z}_\text{c}$ its column-major counterpart, $\overleftarrow{\cdot}$ a
sequence reversal and $W_\text{m}\!\in\!\mathbb{R}^{d\times 4d}$. At $32^2$ and
$16^2$ a single left-to-right scan is used. Decoder stages inherit the setting
of the encoder stage at the same resolution, so four-directional scanning
appears on both sides of the network at $128^2$ and $64^2$ and nowhere else.
This takes the network from 32 scan instances under uniform four-directional
scanning to 20. Section~\ref{sec:scan} reports what that does to accuracy and
cost.

\subsection{Conditional Domain Feature Routing}

HC-18 and PSFHS differ in probe geometry, field of view and tissue
appearance. Rather than train a model per domain or require a scanner label at
test time, CDFR learns a per-domain channel recalibration in the manner of
squeeze-and-excitation~\cite{hu2018senet} and collapses it at inference.

A convolution refines the bottleneck features $\mathbf{x}_b$. Global average
pooling and a two-layer MLP ($C\!\to\!C/4\!\to\!C/4$, $C\!=\!320$) produce an
embedding $\mathbf{s}$, and two domain-specific heads each emit a sigmoid gate
$\mathbf{w}^{(k)}\!\in\![0,1]^{C}$:
\begin{equation}
  \mathbf{x}_\text{out}
    = \text{Conv}(\mathbf{x}_b) \odot \mathbf{w}^{(k)} + \mathbf{x}_b.
  \label{eq:cdfr}
\end{equation}
During training the ground-truth acquisition type selects the active head, so
domain-specific gradients reach only its parameters; no domain-classification
loss is used. At inference the gates are averaged,
$\mathbf{w}_\text{inf}\!=\!\tfrac{1}{2}(\mathbf{w}^{(1)}\!+\!\mathbf{w}^{(2)})$,
and no label is consulted. The two heads do learn different routings: on
held-out images their gate vectors have cosine similarity 0.75 and differ by
0.40 per channel on average.

\subsection{DPAF Decoder}

Each decoder stage upsamples by $2\times$ bilinearly and fuses the result with
the matching encoder skip through a Dual-Path Attention Fusion block.
Attention on skip connections has been used to suppress irrelevant encoder
activations~\cite{oktay2018attention,bhilwarawala2025attres}; DPAF concatenates encoder and decoder
features $\mathbf{f}_\text{cat}$ and applies channel attention followed by
spatial attention, as in CBAM~\cite{woo2018cbam}:
\begin{align}
  \mathbf{f}_c &= \mathbf{f}_\text{cat}
    \odot \sigma\!\bigl(\text{MLP}(\text{GAP}(\mathbf{f}_\text{cat}))\bigr),
    \notag \\
  \mathbf{f}_s &= \mathbf{f}_c
    \odot \sigma\!\bigl(
      \text{Conv}_{7}([\text{Max}(\mathbf{f}_c);\text{Avg}(\mathbf{f}_c)])
    \bigr),
  \label{eq:dpaf}
\end{align}
with an MLP reduction ratio of 8. Four $3{\times}3$ convolutions refine
$\mathbf{f}_s$, followed by a Mamba block at the stage's scan setting, to
attenuate speckle carried along the skip before it reaches the boundary.

\subsection{Training Objective}

\begin{equation}
  \mathcal{L}
    = \mathcal{L}_\text{seg}
    + 0.4\,\mathcal{L}_\text{aux}^{(4)}
    + 0.2\,\mathcal{L}_\text{aux}^{(3)}
    + 0.05\,\mathcal{L}_\text{lm},
  \label{eq:loss}
\end{equation}
where $\mathcal{L}_\text{seg}\!=\!\tfrac{1}{2}\mathcal{L}_\text{focal}\!+\!\tfrac{1}{2}\mathcal{L}_\text{dice}$
combines focal loss~\cite{lin2017focal} ($\gamma\!=\!2$, $\alpha\!=\!0.25$,
label smoothing 0.05) with soft Dice~\cite{milletari2016vnet}. The two
auxiliary terms apply the same composite at decoder stages 4 and 3 as deep
supervision. $\mathcal{L}_\text{lm}$ is a mean squared error on the four
boundary extrema of the mask (topmost, bottommost, leftmost and rightmost
pixel, normalised to $[0,1]$), regressed from decoder stage 4. It acts as a
weak elliptical shape prior. Algorithm~\ref{alg:ff} gives the complete
forward pass and marks where training and inference differ.

\begin{algorithm}[t]
\caption{FETALFusion forward pass, with the training and inference paths}
\label{alg:ff}
\footnotesize
\begin{algorithmic}[1]
\Require image $\mathbf{x}$; scan schedule $\mathcal{D}\!=\!(4,4,1,1)$;
  domain label $k$ (training only)
\Ensure mask $\hat{\mathbf{y}}$; training-only outputs
  $\hat{\mathbf{y}}_4,\hat{\mathbf{y}}_3,\hat{\mathbf{p}}$
\State $\mathbf{u}\gets\mathrm{FullRes}(\mathbf{x})$,\;
  $\mathbf{h}\gets\mathrm{Stem}(\mathbf{x})$
\For{$i=1,\dots,4$} \Comment{encoder}
  \State $\mathbf{h}\gets\mathrm{Proj}_i(\mathbf{h})$
  \State $\mathbf{s}_i\gets\mathrm{CNN}_\mathrm{res}(\mathbf{h})
    +\mathrm{SSM}_{\mathcal{D}_i}(\mathbf{h})$ \Comment{Eqs.~(\ref{eq:fused}), (\ref{eq:4dir})}
  \State $\mathbf{h}\gets\mathrm{Down}(\mathbf{s}_i)$ \Comment{stride-2 conv}
\EndFor
\State $\mathbf{c}\gets\mathrm{Conv}(\mathbf{h})$,\;
  $\mathbf{e}\gets\mathrm{MLP}(\mathrm{GAP}(\mathbf{c}))$
\If{training}
  \State $\mathbf{w}\gets\sigma(g_k(\mathbf{e}))$ \Comment{true domain}
\Else
  \State $\mathbf{w}\gets\tfrac{1}{2}\textstyle\sum_{j=1}^{2}\sigma(g_j(\mathbf{e}))$
    \Comment{no label}
\EndIf
\State $\mathbf{h}\gets\mathbf{c}\odot\mathbf{w}+\mathbf{h}$ \Comment{Eq.~(\ref{eq:cdfr})}
\For{$i=4,\dots,1$} \Comment{decoder}
  \State $\mathbf{h}\gets\mathrm{DPAF}(\mathbf{s}_i,\mathrm{Up}(\mathbf{h}))$
    \Comment{Eq.~(\ref{eq:dpaf})}
  \State $\mathbf{h}\gets\mathrm{SSM}_{\mathcal{D}_i}(\mathrm{Conv}(\mathbf{h}))$
  \State \textbf{if} $i=4$: $\hat{\mathbf{y}}_4\gets\mathrm{Aux}_4(\mathbf{h})$,\;
    $\hat{\mathbf{p}}\gets\mathrm{LM}(\mathbf{h})$
  \State \textbf{if} $i=3$: $\hat{\mathbf{y}}_3\gets\mathrm{Aux}_3(\mathbf{h})$
\EndFor
\State $\hat{\mathbf{y}}\gets\sigma\bigl(\mathrm{Head}(\mathrm{Refine}(
  [\mathrm{Up}(\mathbf{h});\mathbf{u}]))\bigr)$
\State \textbf{if} training: minimise Eq.~(\ref{eq:loss}) \Comment{landmarks from the mask}
\end{algorithmic}
\end{algorithm}

\section{Experiments}

\subsection{Data and Protocol}

\begin{table*}[t]
  \caption{Held-out test set ($n\!=\!471$), mean over images. All methods trained
    under one schedule, from scratch unless stated. FETALFusion uses averaged
    CDFR gates and no domain label, and is the first of two training runs
    (Section~\ref{sec:noise}). HD95 med.: median HD95. $>$10\,mm: images with
    HD95 above 10\,mm. TTA: four-view flip test-time augmentation for every
    method. Latency: one $256{\times}256$ image on a T4. Bold: best among the
    models trained from scratch.}
  \label{tab:main}
  \centering
  \footnotesize
  \setlength{\tabcolsep}{3.3pt}
  \begin{tabular}{lccccccccccc}
    \toprule
    & & & \multicolumn{7}{c}{No test-time augmentation} & \multicolumn{2}{c}{With TTA} \\
    \cmidrule(lr){4-10}\cmidrule(lr){11-12}
    Method & Params (M) & ms/img & DSC$\uparrow$ & IoU$\uparrow$
           & HD95$\downarrow$ & HD95 med.$\downarrow$ & ASD$\downarrow$
           & NSD$\uparrow$ & $>$10\,mm$\downarrow$ & DSC$\uparrow$ & HD95$\downarrow$ \\
    \midrule
    U-Net~\cite{ronneberger2015unet} & 31.0 & 40.6 & .9600 & .9272 & 2.32 & 1.51 & 0.78 & .8816 & 9 & .9608 & 2.10 \\
    Att.\ U-Net~\cite{oktay2018attention} & 31.4 & 40.8 & .9603 & .9272 & 2.26 & 1.51 & 0.78 & .8804 & 9 & .9610 & 2.15 \\
    ResU-Net (nnU-Net-style)~\cite{isensee2021nnu} & 5.9 & 9.4 & .9557 & .9180 & 2.70 & 1.73 & 0.88 & .8556 & 12 & .9592 & 2.42 \\
    ResNet-34 U-Net~\cite{he2016resnet} & 24.4 & 6.4 & .9605 & .9278 & 2.02 & \textbf{1.46} & 0.75 & .8850 & 4 & .9626 & 1.88 \\
    UNet++ (ResNet-34)~\cite{zhou2018unetpp} & 26.1 & 11.9 & \textbf{.9640} & \textbf{.9323} & \textbf{1.88} & 1.48 & \textbf{0.70} & \textbf{.8929} & \textbf{2} & .9645 & \textbf{1.85} \\
    TransUNet-lite~\cite{chen2021transunet} & 18.8 & 37.0 & .9586 & .9226 & 2.42 & 1.74 & 0.82 & .8649 & 7 & .9602 & 2.23 \\
    VM-UNet~\cite{ruan2024vmunet} & 30.0 & 34.5 & .9337 & .8842 & 3.62 & 2.50 & 1.27 & .7609 & 17 & .9373 & 3.09 \\
    U-Mamba~\cite{ma2024umamba} & 7.3 & 11.8 & .9601 & .9253 & 2.41 & 1.63 & 0.81 & .8721 & 8 & .9631 & 2.07 \\
    \midrule
    \textbf{FETALFusion} & 16.8 & 36.6 & .9627 & .9301 & 2.10 & 1.52 & 0.75 & .8855 & 4 & \textbf{.9651} & 1.87 \\
    \midrule
    \multicolumn{12}{l}{\textit{ImageNet initialisation (all rows above are trained from scratch)}} \\
    ResNet-34 U-Net, ImageNet~\cite{he2016resnet} & 24.4 & 6.4 & .9664 & .9365 & 1.75 & 1.40 & 0.66 & .9051 & 2 & .9678 & 1.65 \\
    \bottomrule
  \end{tabular}
\end{table*}

\noindent\textbf{Datasets.}
HC-18~\cite{hc18} provides 999 transabdominal images, mostly
$800{\times}540$ pixels, with elliptical outline annotations, per-image pixel
spacing and a reference HC. Spacing varies widely, from 0.055 to
0.308\,mm/px on our test split (mean 0.143). We close and fill the outlines;
five needed an extra dilation step to close, and every test-set mask reproduces the
reference HC to within 2.5\,mm, so filling does not contribute to test error.
PSFHS~\cite{psfhs} provides 1,358 intrapartum transperineal images at
$256{\times}256$, from which we take the fetal-head class.

\noindent\textbf{Splits.}
Each dataset is split 70/10/20 into training, validation and test,
stratified by acquisition type: 1,650 training, 236 validation and 471 test
images (200 HC-18, 271 PSFHS). Early stopping and checkpoint selection use
the validation split, scored through the label-free inference path. Every
number reported here comes from the test split, which no model sees during
training or selection.

\noindent\textbf{Training.}
One configuration is used for every model in the paper, including both
baselines and all ablation variants: up to 50 epochs with early stopping
(patience 12), AdamW~\cite{loshchilov2019decoupled} (learning rate and weight
decay $10^{-4}$), a 5-epoch linear warm-up followed by cosine decay, gradient
clipping at norm 1.0, batch size 16, full precision, and one augmentation
pipeline (flips, rotation within $\pm20^{\circ}$, elastic deformation with
$\alpha\!=\!120$ and $\sigma\!=\!10$, brightness and contrast jitter, Gaussian
blur, sharpness adjustment). All models are trained from scratch on a single
NVIDIA T4 (16\,GB).

\noindent\textbf{Metrics.}
We report Dice (DSC), IoU, HD95, average surface distance (ASD) and
normalised surface Dice (NSD) at a 1.5\,mm tolerance~\cite{nikolov2021}.
Surface pixels are extracted by binary erosion and distances come from an
exact Euclidean distance transform over all of them, with no subsampling.
Predictions are resampled to native resolution before measurement, so each
HC-18 image is measured with its own spacing. PSFHS provides no spacing; we
apply a nominal 0.28\,mm/px to every method, which makes PSFHS millimetre
values a fixed rescaling of pixel distances. Comparisons are unaffected, but
absolute PSFHS distances should not be read as physical. Methods are compared
with the paired Wilcoxon signed-rank test~\cite{wilcoxon1945} on matched
images, Holm correction~\cite{holm1979} across comparisons, and bootstrap 95\%
intervals on the mean paired difference.

\noindent\textbf{Baselines.}
We compare against two state-space segmenters, U-Mamba~\cite{ma2024umamba}
and VM-UNet~\cite{ruan2024vmunet}; five convolutional ones, U-Net,
Attention U-Net, a residual U-Net with nnU-Net's 2D defaults, and U-Net and
UNet++ with a ResNet-34 encoder (Table~\ref{tab:main}); and TransUNet in a
lite configuration. All are trained from scratch under the protocol above; the
ResNet-34 models use segmentation\_models\_pytorch. Our U-Mamba appends a
Mamba block to each residual encoder stage at $64^2$ and below, and our VM-UNet
uses four-directional selective-scan blocks after a $4{\times}4$ patch
embedding. The published VM-UNet initialises from ImageNet-pretrained VMamba
weights and reports lower accuracy without them~\cite{ruan2024vmunet}, so it
should be read as a weak baseline. To measure what initialisation
contributes, we also train the ResNet-34 U-Net from ImageNet weights, the only
pretrained model in the paper.

\subsection{Comparison with Baselines}

FETALFusion improves on U-Mamba on all five metrics (Table~\ref{tab:main}).
HD95 falls by 0.31\,mm (95\% interval $[-0.57,-0.05]$) and DSC rises by
0.0026 ($[0.0008,0.0043]$), with Holm-corrected $p\le1.3{\times}10^{-5}$ on
every metric. VM-UNet trails both by a wide margin. The lead over U-Mamba
does not depend on the inference protocol: with four-view flip augmentation
applied to every method, FETALFusion still leads by 0.20\,mm in HD95
($[-0.34,-0.11]$) and 0.0020 in DSC ($p\le2.4{\times}10^{-4}$).
Fig.~\ref{fig:hd95}a shows the whole distribution. FETALFusion's curve sits
to the left of U-Mamba's, with a lower HD95 at the 25th, 50th, 75th and 90th percentiles. The second
training run of the full model also leads U-Mamba on all five metrics (HD95
$-0.42$\,mm, every $p<10^{-7}$; Section~\ref{sec:noise}).

By acquisition type the gain over U-Mamba is clearest on PSFHS, where HD95
falls by 0.23\,mm ($[-0.33,-0.12]$) and the median from 1.70 to 1.58\,mm. On
HC-18 the mean falls by 0.43\,mm but the interval crosses zero
($[-1.01,0.15]$) and the medians are level (1.51 against 1.49\,mm). The count of images above 10\,mm (4 against 8) does not survive
augmentation (3 against 2), so we do not read it as a robust difference.

Against the convolutional and transformer baselines the ranking follows the
encoder. FETALFusion improves on U-Net, Attention U-Net, the nnU-Net-style
ResU-Net and TransUNet-lite in mean DSC and HD95 under both protocols, each
by more than twice its run-to-run s.d. A ResNet-34 U-Net trained from scratch
is level with it (DSC 0.0022 lower, HD95 0.08\,mm lower); UNet++ with the same
encoder exceeds it without augmentation (DSC $+0.0013$, HD95 $-0.22$\,mm) and
ties under augmentation; and ImageNet initialisation gives the best result in
Table~\ref{tab:main} (DSC 0.9664, HD95 1.75\,mm) at about a sixth of
FETALFusion's latency. These single runs are compared on aggregates against
FETALFusion's run-to-run variation, not tested for significance. They agree
with the ablation (Section~\ref{sec:ablation}): at this data scale a strong,
ideally pretrained, convolutional encoder matters more than the state-space
branch. The standard FLOP counter misses the fused selective-scan kernel, so we
report latency instead.

\subsection{Label-Free Inference}

Supplying the true acquisition type to the CDFR gates leaves DSC unchanged to
four decimals and lowers HD95 by 0.024\,mm, well inside the run-to-run
variation; with test-time augmentation the difference vanishes (below
0.001\,mm). The averaged gates cost nothing, so label-free deployment is a
property of the method. Test-time augmentation itself adds 0.0024 to DSC and
removes 0.23\,mm from HD95.

\subsection{Run-to-Run Variation}
\label{sec:noise}

The selective-scan kernel and data-loader ordering are not deterministic, so
identical settings give different models. Over two independent training runs
the full model reaches
DSC $0.9629\pm0.0004$, HD95 $2.05\pm0.08$\,mm and NSD $0.886\pm0.001$
(mean$\pm$s.d.); the paired test does not distinguish the runs on any of the
five metrics ($p>0.13$). The variation sits almost entirely in HC-18, where
the runs differ by 0.26\,mm in HD95 against 0.005\,mm on PSFHS, driven by
the small-head failures of Section~\ref{sec:errors}. An ablation effect
counts only if it is significant after Holm correction across the seven
variants, against each run, and larger than the difference between runs.

\subsection{Ablation}
\label{sec:ablation}

\begin{table}[t]
  \caption{Component ablation, held-out test set, no test-time
    augmentation. Full model: mean over two independent training runs, with the s.d.\ in
    the $\Delta$ columns. $\Delta$: relative to that mean. Bold: significant after Holm
    correction against each run and larger than the difference between runs.}
  \label{tab:ablation}
  \centering
  \footnotesize
  \setlength{\tabcolsep}{2.5pt}
  \begin{tabular}{lcccccc}
    \toprule
    Variant & DSC & $\Delta$ & HD95 & $\Delta$ & NSD & $\Delta$ \\
    \midrule
    Full model (2 runs) & .9629 & $\pm$.0004 & 2.047 & $\pm$.076 & .8863 & $\pm$.0012 \\
    \midrule
    w/o SSM branch & .9628 & $-$.0001 & 1.944 & $-$.103 & .8887 & $+$.0024 \\
    w/o DPAF & .9628 & $-$.0001 & 2.135 & $+$.088 & .8871 & $+$.0008 \\
    w/o landmark prior & .9589 & \textbf{$-$.0040} & 2.143 & $+$.096 & .8684 & \textbf{$-$.0179} \\
    Shared gate, no CDFR & .9612 & \textbf{$-$.0017} & 2.093 & $+$.046 & .8777 & $-$.0086 \\
    \bottomrule
  \end{tabular}
\end{table}

Table~\ref{tab:ablation} removes one component at a time, each trained from
scratch under the same schedule. The landmark prior is the clearest contributor: without it DSC falls by
0.0040 and NSD by 0.0179, the largest effects in Table~\ref{tab:ablation},
although the term carries a weight of only 0.05. Its HD95 change
($+0.10$\,mm) is smaller than the difference between runs, so we do not claim
it. Replacing the two CDFR gates with one shared gate lowers pooled DSC by
0.0017, and the effect comes from PSFHS: there the shared gate is worse on all
three metrics (HD95 $+0.08$\,mm with intervals excluding zero, NSD $-0.0145$;
$p\le0.01$ against each run, surviving Holm correction over the six
domain-metric tests), about eighteen times the run-to-run variation on that
domain, while on HC-18 it makes no detectable difference.

Removing the state-space branch leaves DSC unchanged, and its mean HD95 sits
0.103\,mm below the full model, just inside the difference between runs; no
paired test finds a consistent change (Holm-corrected $p\ge0.49$ on every
metric), and it trains in a third of the time (58 against 172 minutes).
Removing DPAF is likewise not significant. At this data scale the
convolutional path already recovers the head.

\subsection{Scan Schedule: Accuracy and Cost}
\label{sec:scan}

\begin{table}[t]
  \caption{Scan directions at $128^2$, $64^2$, $32^2$, $16^2$. Accuracy on the
    held-out test set; cost on one T4. SSM: state-space parameters (M). Mem:
    peak training memory at batch 8 (GB). Ours: mean of the two full-model
    runs.}
  \label{tab:scan}
  \centering
  \footnotesize
  \setlength{\tabcolsep}{3.2pt}
  \begin{tabular}{lcccccc}
    \toprule
    Schedule & DSC & HD95 & NSD & SSM & ms/img & Mem \\
    \midrule
    1-1-1-1 & .9620 & 2.048 & .8777 & 2.44 & 22.0 & 4.64 \\
    4-1-1-1 & .9587 & 2.179 & .8670 & 2.67 & 31.4 & 6.47 \\
    \textbf{4-4-1-1} (ours) & .9629 & 2.047 & .8863 & 3.20 & 35.9 & 7.18 \\
    4-4-4-4 & .9624 & 1.957 & .8797 & 9.20 & 42.4 & 7.63 \\
    \bottomrule
  \end{tabular}
\end{table}

Table~\ref{tab:scan} sweeps where the four-directional scans go, and placement
matters. Restricting them to the highest resolution (4-1-1-1) is
significantly worse than the resolution-aware schedule in DSC ($-0.0042$) and
NSD ($-0.0193$) against each run, although its HD95 change ($+0.13$\,mm) is
not consistent across runs. Extending them to every stage (4-4-4-4) buys
nothing: accuracy is tied, while
the resolution-aware schedule needs 35\% of its state-space parameters, 15\% less latency and 6\% less training memory; the two $128^2$ blocks alone
take 37\% of the forward time.

The single-directional schedule (1-1-1-1) is the strongest alternative: it
ties with the resolution-aware schedule after correction and is cheaper (22.0
against 35.9\,ms). The resolution-aware schedule keeps a small NSD edge
($+0.0086$) whose bootstrap interval excludes zero against each run but which
does not survive Holm correction, so the case for multi-directional scanning
rests on that margin. What the sweep does establish is that four-directional
scans, when used, belong at $128^2$ and $64^2$ together and not below. The dip
at 4-1-1-1 is a single run we cannot explain, so we treat it with care.

\subsection{Where the Errors Are}
\label{sec:errors}

\begin{figure}[t]
  \centering
  \includegraphics[width=\columnwidth]{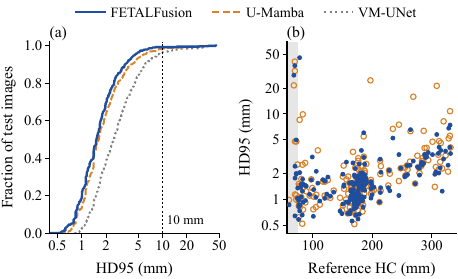}
  \caption{(a) Cumulative distribution of per-image HD95 on the 471 test
    images, no test-time augmentation. (b) HD95 against reference HC on the
    200 HC-18 test images. Filled: FETALFusion; hollow: U-Mamba. The shaded
    band marks heads below the 5th percentile of HC (74\,mm), where the
    largest errors of both methods occur.}
  \label{fig:hd95}
\end{figure}

The largest boundary errors concentrate on the smallest heads
(Fig.~\ref{fig:hd95}b). On the ten HC-18 test images below the 5th
percentile of HC (74\,mm), FETALFusion's mean HD95 is 8.1\,mm against
2.0\,mm on the rest (U-Mamba: 11.1\,mm). Its three largest errors are
early-gestation heads (HC 68 to 78\,mm), over-segmented to 1.4 to 2.1 times
the true area. By DSC, the bottom decile is almost entirely transperineal (43
of 47). Size-aware sampling is the obvious remedy.

\subsection{Clinical HC Agreement}

HC is fitted as an ellipse to the predicted mask at native resolution with
each image's pixel spacing, and compared against the HC-18 reference HC rather
than one re-derived from the ground-truth mask.
On the 200 HC-18 test images FETALFusion reaches Pearson $r\!=\!0.9987$,
ICC(2,1)\,=\,0.9986~\cite{shrout1979}, a mean bias of $+0.44$\,mm, 95\% limits
of agreement of $[-6.59,+7.47]$\,mm~\cite{bland1986} and a mean absolute error
of 2.61\,mm, with 88\% of estimates within 5\,mm. U-Mamba reaches ICC 0.9979,
wider limits ($[-8.20,+9.34]$\,mm) and a mean absolute error of 2.90\,mm; the difference in absolute error is not significant ($p\!=\!0.38$). The
ImageNet-initialised U-Net does better still (ICC 0.9990, mean absolute error
2.35\,mm). For scale, an
ellipse fitted to the ground-truth mask itself has a mean absolute error of
1.15\,mm, the share owed to the measurement pipeline.

\section{Conclusion}

FETALFusion segments the fetal head across two acquisition types with
DSC 0.963, HD95 2.10\,mm and HC agreement at ICC 0.9986, needs no domain label at test time, and improves on the state-space baselines
and on the classic convolutional and transformer baselines under a protocol in
which every model is trained and measured the same way. The domain gates earn
their place on the harder, transperineal domain, and the landmark prior is the
most useful single component.

The ablation is equally informative about what does not drive accuracy at
this scale. Removing the state-space
branch does not measurably change accuracy, and single-directional scanning
ties with the resolution-aware schedule. The latter is best read as the
efficient way to use four-directional scanning: it matches uniform scanning at
a third of the state-space parameters and beats placing the extra directions
at the highest resolution alone. The baseline comparison points the same way from the other side: ResNet-34
encoders match or exceed FETALFusion, and ImageNet initialisation helps most.
Carrying the parts that do help, the landmark prior and the domain gates, onto
a pretrained residual encoder is the natural next step.

We recommend surface distances on a held-out split, with a measured
run-to-run variation, as the default protocol for this task.

\noindent\textbf{Code.}
Training, evaluation and analysis notebooks, the exact splits and the
per-image results behind every table are available at
\url{https://github.com/Ammar-ss/FETALFusion}.

\section*{Limitations}

We use two public datasets and train the full model twice and every other
configuration once; more training runs would sharpen the HD95 comparisons on HC-18,
where run-to-run drift is largest. The baselines are single runs compared on aggregates; all but one are trained
from scratch, which disadvantages VM-UNet and TransUNet in particular, and
FETALFusion is larger than our U-Mamba, so that comparison does not isolate
design from capacity.
PSFHS millimetre figures rest on a nominal pixel spacing, and CDFR is untested
beyond two domains. No cross-scanner, prospective or clinical-workflow
evaluation has been done, and none of these results is evidence of readiness
for clinical use.


\begin{thebibliography}{99}

\bibitem{hc18}
T.~L.~A. van den Heuvel, D.~de Bruijn, C.~L. de Korte, and B.~van Ginneken,
``Automated measurement of fetal head circumference using 2D ultrasound
images,'' \emph{PLoS ONE}, vol.~13, no.~8, p.~e0200412, 2018.

\bibitem{bhilwarawala2025attres}
A.~Bhilwarawala and M.~Bandyopadhyay, ``Attention-ResUNet for automated fetal
head segmentation,'' presented at the 5th Int. Conf. Advanced Network
Technologies and Intelligent Computing (ANTIC), Gwalior, India, Dec. 2025.
arXiv:2604.18148.

\bibitem{chen2021transunet}
J.~Chen, Y.~Lu, Q.~Yu, X.~Luo, E.~Adeli, Y.~Wang, L.~Lu, A.~L. Yuille, and
Y.~Zhou, ``TransUNet: Transformers make strong encoders for medical image
segmentation,'' arXiv:2102.04306, 2021.

\bibitem{gu2024mamba}
A.~Gu and T.~Dao, ``Mamba: Linear-time sequence modeling with selective state
spaces,'' arXiv:2312.00752, 2023.

\bibitem{ma2024umamba}
J.~Ma, F.~Li, and B.~Wang, ``U-Mamba: Enhancing long-range dependency for
biomedical image segmentation,'' arXiv:2401.04722, 2024.

\bibitem{ruan2024vmunet}
J.~Ruan, J.~Li, and S.~Xiang, ``VM-UNet: Vision Mamba UNet for medical image
segmentation,'' arXiv:2402.02491, 2024.

\bibitem{wu2025hvmunet}
R.~Wu, Y.~Liu, P.~Liang, and Q.~Chang, ``H-vmunet: High-order Vision Mamba
UNet for medical image segmentation,'' \emph{Neurocomputing}, vol.~624,
p.~129447, 2025.

\bibitem{zhu2025mcc}
Y.~Zhu, D.~Zhang, Y.~Lin, Y.~Feng, and J.~Tang, ``Merging context clustering
with visual state space models for medical image segmentation,'' \emph{IEEE
Trans. Med. Imag.}, vol.~44, no.~5, pp.~2131--2142, 2025.

\bibitem{ronneberger2015unet}
O.~Ronneberger, P.~Fischer, and T.~Brox, ``U-Net: Convolutional networks for
biomedical image segmentation,'' in \emph{Proc. MICCAI}, 2015, pp.~234--241.

\bibitem{ulyanov2016instance}
D.~Ulyanov, A.~Vedaldi, and V.~Lempitsky, ``Instance normalization: The
missing ingredient for fast stylization,'' arXiv:1607.08022, 2016.

\bibitem{isensee2021nnu}
F.~Isensee, P.~F. Jaeger, S.~A.~A. Kohl, J.~Petersen, and K.~H. Maier-Hein,
``nnU-Net: A self-configuring method for deep learning-based biomedical image
segmentation,'' \emph{Nature Methods}, vol.~18, no.~2, pp.~203--211, 2021.

\bibitem{hu2018senet}
J.~Hu, L.~Shen, and G.~Sun, ``Squeeze-and-excitation networks,'' in
\emph{Proc. CVPR}, 2018, pp.~7132--7141.

\bibitem{oktay2018attention}
O.~Oktay, J.~Schlemper, L.~Le~Folgoc, M.~Lee, M.~Heinrich, K.~Misawa, K.~Mori,
S.~McDonagh, N.~Y. Hammerla, B.~Kainz, B.~Glocker, and D.~Rueckert,
``Attention U-Net: Learning where to look for the pancreas,''
arXiv:1804.03999, 2018.

\bibitem{woo2018cbam}
S.~Woo, J.~Park, J.-Y. Lee, and I.~S. Kweon, ``CBAM: Convolutional block
attention module,'' in \emph{Proc. ECCV}, 2018, pp.~3--19.

\bibitem{lin2017focal}
T.-Y. Lin, P.~Goyal, R.~Girshick, K.~He, and P.~Doll\'ar, ``Focal loss for
dense object detection,'' in \emph{Proc. ICCV}, 2017, pp.~2980--2988.

\bibitem{milletari2016vnet}
F.~Milletari, N.~Navab, and S.-A. Ahmadi, ``V-Net: Fully convolutional neural
networks for volumetric medical image segmentation,'' in \emph{Proc. 3DV},
2016, pp.~565--571.

\bibitem{he2016resnet}
K.~He, X.~Zhang, S.~Ren, and J.~Sun, ``Deep residual learning for image
recognition,'' in \emph{Proc. CVPR}, 2016, pp.~770--778.

\bibitem{zhou2018unetpp}
Z.~Zhou, M.~M. Rahman Siddiquee, N.~Tajbakhsh, and J.~Liang, ``UNet++: A nested
U-Net architecture for medical image segmentation,'' in \emph{Proc. DLMIA,
MICCAI Workshops}, LNCS vol.~11045, 2018, pp.~3--11.

\bibitem{psfhs}
G.~Chen, J.~Bai, Z.~Ou, Y.~Lu, and H.~Wang, ``PSFHS: Intrapartum ultrasound
image dataset for AI-based segmentation of pubic symphysis and fetal head,''
\emph{Scientific Data}, vol.~11, no.~1, p.~436, 2024.

\bibitem{loshchilov2019decoupled}
I.~Loshchilov and F.~Hutter, ``Decoupled weight decay regularization,'' in
\emph{Proc. ICLR}, 2019.

\bibitem{nikolov2021}
S.~Nikolov \emph{et al.}, ``Clinically applicable segmentation of head and
neck anatomy for radiotherapy: Deep learning algorithm development and
validation study,'' \emph{J. Med. Internet Res.}, vol.~23, no.~7,
p.~e26151, 2021.

\bibitem{wilcoxon1945}
F.~Wilcoxon, ``Individual comparisons by ranking methods,'' \emph{Biometrics
Bulletin}, vol.~1, no.~6, pp.~80--83, 1945.

\bibitem{holm1979}
S.~Holm, ``A simple sequentially rejective multiple test procedure,''
\emph{Scand. J. Statist.}, vol.~6, no.~2, pp.~65--70, 1979.

\bibitem{shrout1979}
P.~E. Shrout and J.~L. Fleiss, ``Intraclass correlations: Uses in assessing
rater reliability,'' \emph{Psychological Bulletin}, vol.~86, no.~2,
pp.~420--428, 1979.

\bibitem{bland1986}
J.~M. Bland and D.~G. Altman, ``Statistical methods for assessing agreement
between two methods of clinical measurement,'' \emph{The Lancet},
vol.~327, no.~8476, pp.~307--310, 1986.

\end{thebibliography}
\end{document}